\documentclass[runningheads]{llncs}
\usepackage{graphicx}
\usepackage{amsmath,graphicx}

\usepackage{enumitem}
\setlist{nosep, leftmargin=14pt}

\usepackage{mwe} % to get dummy images

\usepackage[usenames,dvipsnames,table]{xcolor}
\definecolor{additionalblue}{RGB}{0,51,102}

\usepackage{tikz}
\usetikzlibrary{positioning,shapes,patterns,calc,backgrounds}
\usepackage{pgfplots}
\usepgfplotslibrary{fillbetween}

\pgfplotsset{
legend image code/.code={
\draw[mark repeat=2,mark phase=2]
plot coordinates {
(0cm,0cm)
(0.15cm,0cm)
(0.3cm,0cm)
};%
}
}

\usepackage{tikzpagenodes}

\usepackage{makecell}
\usepackage{hhline}
\usepackage{multirow}
\newcolumntype{L}[1]{>{\raggedright\let\newline\\\arraybackslash\hspace{0pt}}m{#1}}
\newcolumntype{C}[1]{>{\centering\let\newline\\\arraybackslash\hspace{0pt}}m{#1}}
\newcolumntype{R}[1]{>{\raggedleft\let\newline\\\arraybackslash\hspace{0pt}}m{#1}}
\usepackage{booktabs}

\definecolor{myblue}{RGB}{56,93,138}
\definecolor{myred}{RGB}{192,80,77}
\definecolor{myredlight}{RGB}{223,165,164}
\definecolor{mygreen}{RGB}{155,187,89}
\definecolor{mygreenlight}{RGB}{215,227,188}

\usepackage{amssymb}
\usepackage{bm}

\usepackage{calc}
\usepackage[hidelinks]{hyperref}
\hypersetup{
    pdftitle={Mitigating Unknown Bias in Deep Learning Based Assessment of CT Images},
    pdfauthor={{Simon Langer}, {Oliver Taubmann}, {Felix Denzinger}, {Andreas Maier}, {Alexander M\"uhlberg}},
    pdfsubject={Mitigating Unknown Bias in Deep Learning Based Assessment of CT Images},
    pdfkeywords={DL} {Unknown Bias} {Debiasing} {CT} {Technical Variation},
}

\usepackage{dblfloatfix} % for bottom of page

\usepackage{soul}
\usepackage{tikz}
\usetikzlibrary{calc}

\usepackage{censor}
\usepackage[textsize=footnotesize,textwidth=1.5cm]{todonotes}
\let\oldtodo\todo
\renewcommand{\todo}[1]{\oldtodo{#1}\PackageWarning{TODO:}{TODO: #1}}

\tikzstyle{notestyleraw} = [
draw=\@todonotes@currentbordercolor,
fill=\@todonotes@currentbackgroundcolor,
text=\@todonotes@currenttextcolor,
line width=0.5pt,
text width = \@todonotes@textwidth - 1.6 ex - 1pt    +1pt,
inner sep = 0.2 ex,
rounded corners=0pt]

\makeatletter
\newif\if@anonymize

\@anonymizetrue
\@anonymizefalse

\if@anonymize
\newcommand{\anonymize}[3][c]{\makebox[0cm][l]{}\makebox[\widthof{#2}][#1]{*}\makebox[0cm][r]{}}

\newcommand{\anonymizedyn}[2]{#2}

\else
  \newcommand{\anonymize}[3][c]{#2}
  
  \newcommand{\anonymizedyn}[2]{#1}
  
\fi

\makeatother

\newcommand{\eg}{e.\,g.}
\newcommand{\ie}{i.\,e.}

\newcommand{\wrt}{w.\,r.\,t.}

\newcommand{\myuncheck}{\begin{tikzpicture}[square/.style={regular polygon,regular polygon sides=4,inner sep=2pt,ultra thick}]%
  \node at (0,0) [square,draw] (s) {};%
\end{tikzpicture}}
\newcommand{\mycheck}{\begin{tikzpicture}[square/.style={regular polygon,regular polygon sides=4,inner sep=2pt,ultra thick}]%
  \node at (0,0) [square,draw,fill] (s) {};%
\end{tikzpicture}}

\begin{document}
\title{DeepTechnome: Mitigating Unknown Bias in Deep Learning Based Assessment of CT Images}
\titlerunning{DeepTechnome}
\author{\anonymizedyn{Simon Langer}{First Author}$^{1}$ \and \anonymizedyn{Oliver Taubmann}{Second Author}$^{2}$ \and \anonymizedyn{Felix Denzinger}{Third Author}$^{1,2}$ \and \anonymizedyn{Andreas Maier}{Fourth Author}$^{1}$ \and \anonymizedyn{Alexander M\"uhlberg}{\\Fifth Author}$^{2}$}
% \author{Anonymous~\\~}%
%
\authorrunning{\anonymizedyn{S. Langer et al.}{Anonymous et al.}}
\institute{
\anonymizedyn{Pattern Recognition Lab, Friedrich-Alexander-Universit\"at Erlangen-N\"urnberg, Germany}{Institution 1\\~}\\
\email{\anonymizedyn{\{simon.langer, felix.denzinger, andreas.maier\}@fau.de}{\{abc,def\}@domain.com}}\\% 
\and
\anonymizedyn{Siemens Healthcare GmbH, Forchheim, Germany}{Institution 2}\\
\email{\anonymizedyn{oliver.taubmann@siemens-healthineers.com, alexander-muehlberg@hotmail.com}{\{abc,def\}@domain2.com}}
}
%
% \institute{
% Anonymous Organization\\~\\
% \email{**@******.***}\\
% % ~\\
% % \and
% ~\\
% \email{~}}

%
\maketitle              % typeset the header of the contribution
\setcounter{footnote}{0} %
\begin{abstract}
Reliably detecting diseases using relevant biological information is crucial for real-world applicability of deep learning techniques in medical imaging.  We debias deep learning models during training against unknown bias -- without preprocessing/filtering the input beforehand or assuming specific knowledge about its distribution or precise nature in the dataset. We use control regions as surrogates that carry information regarding the bias, employ the classifier model to extract features, and suppress biased intermediate features with our custom, modular \mbox{\textit{DecorreLayer}}. We evaluate our method on a dataset of 952 lung computed tomography scans by introducing simulated biases \wrt~reconstruction kernel and noise level and propose including an adversarial test set in evaluations of bias reduction techniques. In a moderately sized model architecture, applying the proposed method to learn from data exhibiting a strong bias, it near-perfectly recovers the classification performance observed when training with corresponding unbiased data.
\end{abstract}
\section{Introduction}
\label{sec:intro}

Similarly to the batch effect in genomics \cite{Leek2010TacklingTW}, technical variation in CT scans occurs for a variety of reasons, becoming especially problematic when it is correlated with the predictive task, for instance due to prior knowledge of the clinician and/or patient of a likely diagnosis, or site-specific differences in patient selection and acquisition protocols within multi-center data sets \cite{Mhlberg2020TheT}.

The range of reconstruction and scan parameters affects the amount and appearance of technical variation present \cite{Diwakar2018ARO,Maier2018IntroductionM}: choosing an appropriate reconstruction kernel forces a tradeoff between detail and noise level; tube current and voltage affect the amount of noise and radiation dose. The severity and presence of beam hardening, scatter, metal, motion and truncation artifacts are influenced by such choices as well as by patient behavior and physiology.

When automating image analysis with deep/machine learning (DL/ML), bias poses a fundamental challenge -- consequently, there has been substantial research and industry interest in it.
When tackling known bias, normalizing the input data \wrt~technical variation in a preprocessing step is a classical approach; a recent example employs DL to preprocess images, converting their style using convolutional neural networks which predict the image differences of reconstruction kernels \cite{Choe2019DeepLI}. Alternatively, adversarial debiasing can be performed, \ie~a secondary network debiases while training using gradient reversal and/or min-max based approaches \cite{Alvi2018TurningAB,Kim2019LearningNT,Zhang2018MitigatingUB}.
In contrast, there has been relatively little work in DL tackling the more general, universal case of unknown bias. Notably, a resampling preprocessing approach accounts for less frequent permutations of image properties -- acquired with a variational autoencoder (VAE) \cite{Kingma2014AutoEncodingVB} -- by sampling the associated images more frequently \cite{Amini2019UncoveringAM}.
In classical ML, RAVEL regresses unwanted features per voxel using control regions \cite{Fortin2016RemovingIT} and the Technome \cite{Mhlberg2020TheT} combines debiasing and training in order to avoid the risk of removing informative biological information when stabilizing during preprocessing. Our work aims to transfer such ideas to the field of DL.

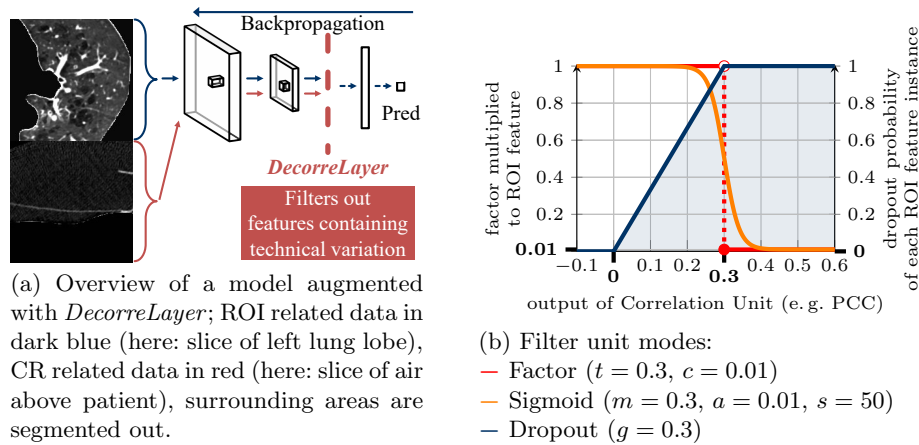
\begin{figure}[b!]
  \centering
  \begin{minipage}[b]{0.45\linewidth}
  \includegraphics[width=1.0\linewidth]{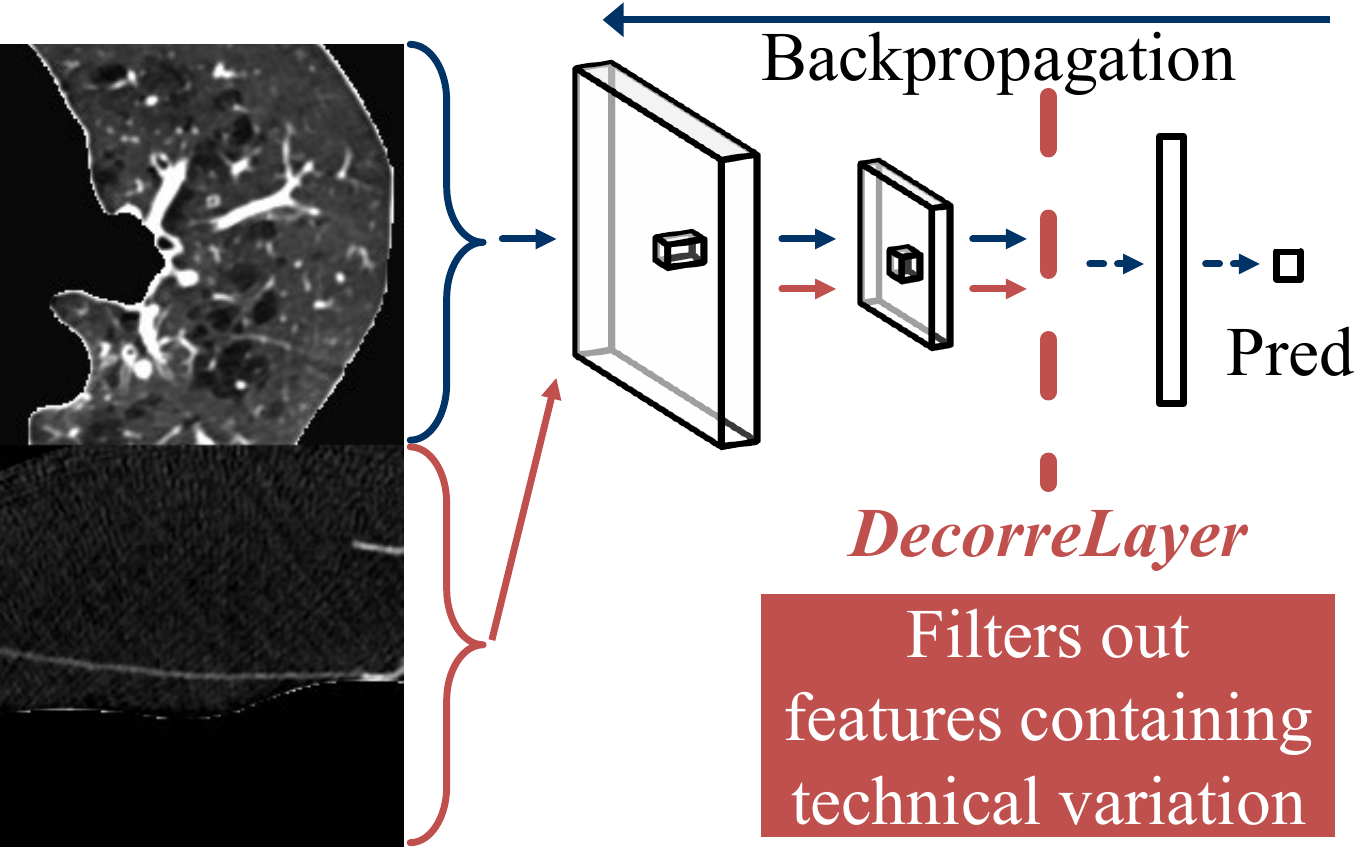}
  \footnotesize(a) Overview of a model augmented with \textit{DecorreLayer}; ROI related data in dark blue (here: slice of left lung lobe), CR related data in red (here: slice of air above patient), surrounding areas are segmented out.
  \end{minipage}%
  \begin{minipage}[b]{0.06\linewidth}
  $~$
  \end{minipage}%
  \begin{minipage}[b]{0.54\linewidth}
      \scriptsize
      \newcommand{\argboldsubscr}[2][\text{roi}]{\ensuremath{\mathbf{#2}_{#1}}}
      \newcommand{\xroi}{\argboldsubscr{X}}
      \newcommand{\xcr}{\argboldsubscr[\text{cr}]{X}}
      \newcommand{\yroi}{\argboldsubscr{\hat{Y}}}
      \newcommand{\dcorr}{\argboldsubscr[f]{\hat{d}}}
      \begin{tikzpicture}
        \newcommand{\plotxmin}{-0.1}
        \newcommand{\plotxmax}{0.6}
        \newcommand{\plotyspace}{0.01}
        \newcommand{\ploty}{4.92848411cm}
        \newcommand{\plotscale}{0.50}
        \newcommand{\plotlegenddist}{0.35}
        \newcommand{\plotrowsep}{0.1cm}
        \begin{axis}[
            opacity=1,
            scale=\plotscale,
            xmin=\plotxmin,
            xmax=\plotxmax,
            ymin=0,
            ymax=1+\plotyspace,
            y=\ploty,
            axis x line*=bottom,
            axis y line=left,
            grid = major,
            ylabel style={align=center, at={(axis cs:-0.05,0.5)}},
            ylabel = {factor multiplied \\to ROI feature},
            extra y ticks={0.01},
            extra tick style={grid=major,font=\boldmath,log identify minor tick positions=false,tick style = {black, ultra thick},tick align = outside},
            xtick=\empty,
            yticklabel style = {
                /pgf/number format/fixed,
                /pgf/number format/precision = 2,
              },
            yticklabel={\ifdim\tick pt=0pt ~ \else\pgfmathprintnumber{\tick}\fi},
            xticklabel={\ifdim\tick pt=0pt ~ \else\pgfmathprintnumber{\tick}\fi},
            scaled x ticks = false,
            legend cell align=left,
            legend style={row sep=0cm,at={(axis cs:1,1)},anchor=north west,draw=none,fill=none,inner xsep=0pt,inner ysep=2pt,nodes={inner sep=2pt,text depth=0.15em},},
          ]
          \addplot[domain=\plotxmin:0.3,red,ultra thick] {1};
          \addplot[domain=0.3:\plotxmax,red,ultra thick,forget plot] {0.01};
          \draw[dotted,red,ultra thick] (axis cs:0.3,1) -- (axis cs:0.3,0.01);
          \pgfplotsset{soldot/.style={color=red,only marks,mark=*}};
          \pgfplotsset{holdot/.style={color=red,fill=white,only marks,mark=*}};
          \addplot[holdot,forget plot] coordinates{(0.3,1)};
          \addplot[soldot,forget plot] coordinates{(0.3,0.01)};
    
          \addplot[
            domain=\plotxmin:\plotxmax,
            samples=1000,
            color=orange,
            ultra thick,
          ]
          {(1-0.01)/(1+e^(50*(x-0.3)))+0.01};
    
        \end{axis}%
        \begin{axis}[
            opacity=1,
            scale=\plotscale,
            xmin=\plotxmin,
            xmax=\plotxmax,
            ymin=0,
            ymax=1+\plotyspace,
            y=\ploty,
            axis x line*=bottom,
            axis y line=right,
            grid = major,
            xlabel = {output of Correlation Unit (\eg~PCC)},
            x label style={at={(axis description cs:0.5,0.02)},anchor=north},
            xtick={-1. , -0.9, -0.8, -0.7, -0.6, -0.5, -0.4, -0.3, -0.2, -0.1,  0. ,  0.1,  0.2,  0.4,  0.5,  0.6,  0.7,  0.8,  0.9,  1.},
            extra x ticks={0,0.3},
            xticklabel={\ifdim\tick pt=0pt ~ \else\pgfmathprintnumber{\tick}\fi},
            extra x tick labels={$\mathbf{0}$,$\mathbf{0.3}$},
            ylabel style={align=center, at={(axis cs:0.53,0.5)}},
            ylabel = {$~$\\dropout probability \\of each ROI feature instance},
            extra y ticks={0},
            extra tick style={grid=major,font=\boldmath,log identify minor tick positions=false,tick style = {black, ultra thick},tick align = outside},
            extra y tick labels={$\mathbf{0}$},
            yticklabel={\ifdim\tick pt=0pt ~ \else\pgfmathprintnumber{\tick}\fi},
            legend cell align=left,
            legend style={row sep=0cm,at={(axis cs:1,0.2)},anchor=north west,draw=none,fill=none,inner xsep=0pt,inner ysep=2pt,nodes={inner sep=2pt,text depth=0.15em},},
          ]
    
          \addplot[name path=f,domain=\plotxmin:\plotxmax,samples=1000,ultra thick,additionalblue] {x < 0 ? 0 : (x > 0.3 ? 1 : x/0.3)};
    
          \path[name path=axis] (axis cs:0,0) -- (axis cs:1,0);
    
          \addplot [
            ultra thick,
            color=additionalblue,
            fill=additionalblue,
            fill opacity=0.1
          ]
          fill between[
              of=f and axis,
              soft clip={domain=0:1},
            ];
        \end{axis}
      \end{tikzpicture}
      $~$\\\footnotesize
      (b) Filter unit modes: \\
      \color{red}$\boldsymbol{-}$\color{black}~Factor ($t=0.3$, $c=0.01$)\\
      \color{orange}$\boldsymbol{-}$\color{black}~Sigmoid ($m=0.3$, $a=0.01$, $s=50$)\\
      \color{additionalblue}$\boldsymbol{-}$\color{black}~Dropout ($g=0.3$)
  \end{minipage}
  
  \caption{\textit{DecorreLayer} overview (a) and details on its Filter Unit module (b).}
  \label{fig:overview}
\end{figure}

\section{Method}
\label{sec:format}
Control regions (CRs) serve as surrogates, capturing the technical variation but little to no biological variation related to the detection task. To maximize similarity \eg~in streak and noise patterns, areas of the CT scan which are outside of, but close to the regions of interest (ROIs) such as surrounding air or anatomical structures are good candidates; the designer chooses depending on the presumed general types of biases that might be present. This introduces the assumption that the confounder manifests not only in the task-relevant area, yet no other knowledge about the specific nature or distribution of the bias is required.

Our novel, modular \textit{DecorreLayer} is inserted into an existing classifier model, receiving features calculated by the model architecture up to that point for both the ROI and CR (applying the same computations to ROI and CR data), and returns a filtered version of the ROI features, as illustrated in Fig.~\ref{fig:overview}a.

\subsection{Correlation Unit}

\textit{DecorreLayer} consists of two modules with separate tasks. The Correlation Unit determines what features probably depend on technical variation, comparing each feature with its control region counterpart over the batch dimension. For the sake of simplicity, the Pearson correlation coefficient (PCC) was used for this purpose in our experiments. In order to operate independently of the location of \eg~a noise pattern, a global average pooled virtual feature vector (one scalar feature as the mean over each pixel in a channel) is used if \textit{DecorreLayer}'s output will still be interpreted with spatial information (\eg~by a subsequent convolutional layer). While the use of PCC may appear to limit the approach to linear co-dependencies, note that it is computed on internal feature maps that are themselves non-linear \wrt{} the input data. Nonetheless, it could also be replaced with other correlation measures, linear or non-linear, if deemed appropriate -- even trainable ones. We suggest inserting \textit{DecorreLayer} prior to every fully connected and convolutional layer, except the first one.

\subsection{Filter Unit}
The Filter Unit is in direct contact with the surrounding architecture. It generates filtered versions of the ROI features as \textit{DecorreLayer}'s output, with stronger filtering being applied when the Correlation Unit's output is higher (Fig.~\ref{fig:overview}b): Either by multiplying with a small constant $c$ if the PCC is above a threshold $t$ (\textit{Factor} mode), a smoothed variant with middle $m$, minimum $a$ and steepness $s$ (\textit{Sigmoid} mode), or by interpreting dependency on technical variation as a dropout probability, guaranteeing dropout at $\geq g$ (\textit{Dropout} mode). The latter performed best, gradually reducing the reliability of a feature the more it is probably dependent on technical variation by increasing the dropout probability appropriately. With $\mathrm{Bern}(p)$ being the Bernoulli distribution, $\mathbf{x}$ denoting all instances of one feature over the batch dimension and $\hat{d}$ the scalar PCC, the result $\mathbf{\hat{y}}$ can be described as\footnote{For the sake of readability, we use a mix of regular and random variables.}:
\begin{align}
  \mathbf{z}                                  \thicksim & ~ \mathrm{Bern} \left( \max\left\{0, 1 - \max\left\{0, \hat{d} \right\} \cdot g^{-1} \right\} \right) \\
  \mathbf{\hat{y}}  =                                   & ~ \mathbf{x} \cdot \mathbf{z}.
\end{align}

\newpage
\subsection{Backward Pass and Inference}
We intentionally relax the constraint of perfectly matching the forward pass computations during the backward pass: \textit{DecorreLayer} is ignored entirely, which results in a reweighting of the error tensor such that errors caused by \textit{DecorreLayer}'s filtering of features containing technical variation are ``blamed on'' the previous layers.
At inference time, \textit{DecorreLayer} is inactive since the model has already learned to extract features without bias, \ie~CRs are no longer needed.

\section{Data Preparation and Testing Setup}
\label{sec:Testing Setup and Data Preparation}
Our evaluation data set contains 952 lung CT scans from a single site, acquired with a \anonymize{SOMATOM\textsuperscript{\textregistered} Force}{Scanner Name} and labeled for the presence (label \texttt{CE}) or absence (\texttt{nCE}) of centrilobular emphysema.
As this disease is primarily visible in the upper half of the lung \cite{Anderson1973CentrilobularEA}, we extract 5 evenly spaced lung-masked axial slices from that area as ROIs, and unrelated air regions above the patient from the same slices as CRs (cf.~Fig.~\ref{fig:overview}a).
Since each scan was reconstructed with both a soft (\anonymize{Br36d3}{type}) and a sharp (\anonymize{Bl57d3}{type}) kernel, we can simulate realistic technical variation by more frequently sampling softer images for label \texttt{CE}, and otherwise sharper, but noisier images (\texttt{CE} $\leftrightarrow$ $90\%$ chance of soft image). In a separate experiment with artificial technical variation we apply additive white Gaussian noise (AWGN, $\mu=0$, $\sigma=0.5\,\sigma_{\text{ROI}}$, $\sigma_{\text{ROI}}$ denotes the standard deviation of the ROI voxel intensities) to images labeled \texttt{CE} with a $90\%$ chance.

We define \textit{adversarial test sets} as an inversion of the introduced manipulations, affecting every image in the test set, and argue for their adoption in future work on debiasing DL/ML since they directly display worst-bias-case performance as well as visualize the reliance of a model on technical variation (such as extra noise) not present in the original, unbiased data (\textit{full test set}).

\begin{figure}[!htb]
  \centering
  \includegraphics[width=\linewidth]{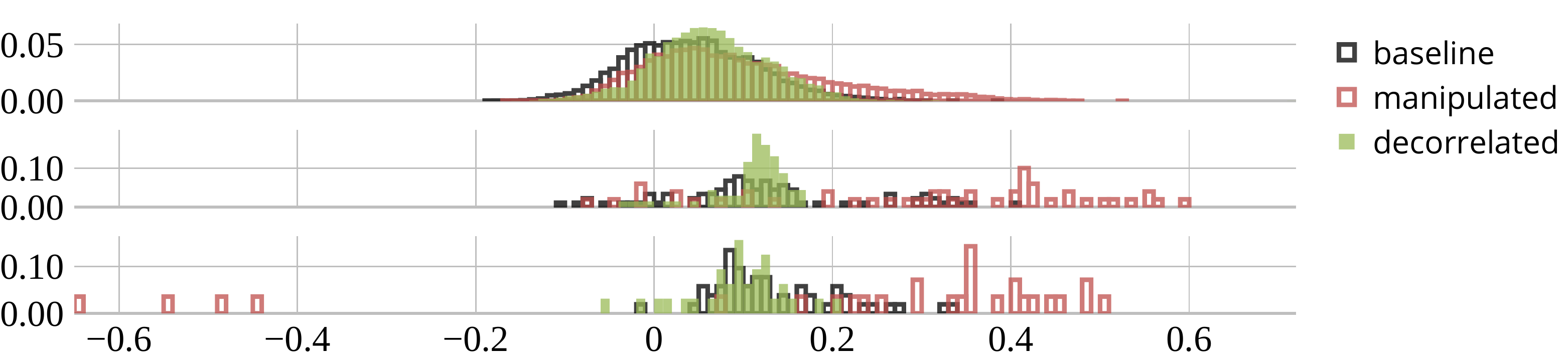}
  \caption{Histogram of Correlations with 3 \textit{DecorreLayer}s inserted at depths 6, 9, 11 (their individual histograms shown top to bottom) into the Small Custom Architecture. We visualize how many activations exhibit a specific (here: Pearson) correlation value (bins on x-axis), normalized by total number of activations (y-axis). Note the shift towards positive correlation when training a model on manipulated, \ie~biased data, compared to the baseline -- \textit{DecorreLayer} successfully reverts this effect.}
  \label{fig:hoc}
\end{figure}

We also propose the Histogram of Correlations -- a visualization technique which plots the correlations of ROI and CR activations (\ie~the Correlation Unit output), optionally including a comparison with unbiased and/or debiased trainings (cf.~Fig.~\ref{fig:hoc}). It both facilitates finding a reasonable hyperparameter range for \textit{DecorreLayer}, and investigating our or other debiasing techniques.

Our approach is tested on three architectures: \textsc{a)} a small custom model consisting of 2 sets of convolutional, ReLU, MaxPool layers (6, 16 channels) followed by 3 fully connected ones, \textsc{b)} a medium custom model consisting of 3 wider sets of this structure (32, 64, 128 channels), followed by 4 fully connected layers, as well as \textsc{c)} ResNet-18 \cite{He2016DeepRL} with batch normalization disabled; training is performed with minimal augmentations (horizontal flip, translation) using stochastic gradient descent.

\begin{figure}[!htb]
  \centering
  \footnotesize
  \newcommand{\plotscalex}{0.4*1.1}
  \newcommand{\plotscaley}{0.6*0.6*0.9}
  \newcommand{\plotscaledist}{0.6*1.1*0.3}
  \begin{tikzpicture}[auto,node distance=3cm,thick,
      mainnode/.style={circle,draw,fill=black,text=white,text width={0.9cm*\plotscalex},align=center},
      mainnodeellipse/.style={ellipse,draw=white,fill=black,text=white,inner sep=0pt,minimum size={1.1cm*\plotscalex},text width={1.2cm*\plotscalex},align=center},
      bgnode/.style={circle,draw=gray,fill=gray,text=white,text width={0.9cm*\plotscalex},align=center},
      bgnodeellipse/.style={ellipse,draw=white,fill=gray,text=white,inner sep=0pt,minimum size={1.1cm*\plotscalex},text width={1.2cm*\plotscalex},align=center},]

    \tikzstyle myBG=[line width=3pt,opacity=1.0]
    \newcommand{\drawLinewithBG}[3]
    {
      \draw[white,myBG]  (#1) -- (#2);
      \draw[black,#3] (#1) -- (#2);
    }

    \node[mainnode, label={[white]center:$S_1$}] (S1) {};
    \node[bgnode, label={[white]center:$S_2$}] (S2) [above right = 1cm*\plotscaledist of S1] {};
    \node[mainnode, label={[white]center:$S_3$}] (S3) [below = 4cm*\plotscaley of S1] {};
    \node[bgnode, label={[white]center:$S_4$}] (S4) [below = 4cm*\plotscaley of S2] {};

    \node[mainnode, label={[white]center:$M_1$}] (M1) [right = 6cm*\plotscalex of S1] {};
    \node[bgnode, label={[white]center:$M_2$}] (M2) [above right = 1cm*\plotscaledist of M1] {};
    \node[mainnode, label={[white]center:$M_3$}] (M3) [below = 4cm*\plotscaley of M1] {};
    \node[bgnode, label={[white]center:$M_4$}] (M4) [below = 4cm*\plotscaley of M2] {};

    \drawLinewithBG{S1}{M1}{dashed};
    \drawLinewithBG{S2}{M2}{dashed};
    \drawLinewithBG{S3}{M3}{dashed};
    \drawLinewithBG{S4}{M4}{dashed};

    \node[] (Divider1) [right = 3cm*\plotscalex of S1] {};
    \node[] (Divider2) [right = 3cm*\plotscalex of S2] {};
    \node[] (Divider3) [right = 3cm*\plotscalex of S3] {};
    \node[] (Divider4) [right = 3cm*\plotscalex of S4] {};

    \path[fill=white, opacity=1, line join=round,line cap=round,miter limit=10.00]
    (Divider1.center) -- (Divider2.center) -- (Divider4.center) -- (Divider3.center) -- (Divider1.center)
    ;
    \path[pattern=crosshatch dots, opacity=1, line join=round,line cap=round,miter limit=10.00]
    (Divider1.center) -- (Divider2.center) -- (Divider4.center) -- (Divider3.center) -- (Divider1.center)
    ;

    \node[mainnode, label={[white]center:$M_1$}] (ExtraM1) [right = 6cm*\plotscalex of S1] {};

    \path[every node/.style={}]
    (S1) edge node [] {} (S2)
    (S1) edge node [] {} (S3)
    (S3) edge node [] {} (S4)
    (S2) edge node [] {} (S4)

    (M1) edge node [] {} (M2)
    (M1) edge node [] {} (M3)
    (M3) edge node [] {} (M4)
    (M2) edge node [] {} (M4);

    \drawLinewithBG{M1}{M3}{};

    \node[] (Origin) [below left = 1cm and 0.8cm of S3] {};
    \node[] (ModelAxisRight) [right = 14cm*\plotscalex of Origin] {Model};
    \node[] (ModelAxisUp) [above = 12.7cm*\plotscaley of Origin] {Manipulation};
    \node[circle] (ModelAxisUpRight) [above right = 1.5cm*\plotscalex of Origin] {};
    \node[] () [below right = 0cm and -0.9cm*\plotscalex*0.6 of ModelAxisUpRight] {Mode: $\bullet$ \textit{Dropout}~~~~\textcolor{gray}{$\bullet$ \textit{Factor}}~~~~\textcolor{gray}{$\tilde{\bullet}$ \textit{Sigmoid}}};

    \node[] (SmallCustom) [below = 1.1cm of S3] {Small Custom};
    \node[] (MediumCustom) [below = 1.1cm of M3] {Medium Custom};

    \node[label={[label distance=-1.7cm,rotate=90,text width=3cm, yshift=0.7cm,align=center]right:Reconstruction\\Kernel}] (ReconstructionKernel) [left = 0.6cm of S1] {};
    \node[label={[label distance=-1.7cm,rotate=90,text width=3cm, yshift=0.7cm,align=center]right:Gaussian\\Noise}] (GaussianNoise) [left = 0.6cm of S3] {};

    \path[every node/.style={}]
    (Origin.center) edge [-latex] node [] {} (ModelAxisRight)
    (Origin.center) edge [-latex] node [] {} (ModelAxisUp)
    (Origin.center) edge [-latex] node [] {} (ModelAxisUpRight)
    ;

    \draw[black,dashed] (S1) -- (M1);

    \node[mainnodeellipse, label={[white]center:$S_1'$}] (S1') [below = -0.3cm*\plotscaley of S1] {};
    \node[mainnodeellipse, label={[white]center:$S_3'$}] (S3') [below = -0.3cm*\plotscaley of S3] {};
    \node[mainnodeellipse, label={[white]center:$M_1'$}] (M1') [below = -0.3cm*\plotscaley of M1] {};
    \node[mainnodeellipse, label={[white]center:$M_3'$}] (M3') [below = -0.3cm*\plotscaley of M3] {};

    \node[bgnodeellipse] (S2t) [above right = -0.3cm*\plotscaley of S2] {$\tilde{S}_2$};

    \node[mainnodeellipse] (S1h) [right = 0.2cm of S1] {${\hat S}_1$};
    \node[mainnodeellipse] (M1h) [right = 0.2cm of M1] {${\hat M}_1$};
    \drawLinewithBG{M1}{M1h}{dashed};
  \end{tikzpicture}
  
  \caption[]{Visualization of core evaluations of our method.
    We keep hyperparameters of nodes connected with solid lines equal (except for the different Filter Unit modes). When fully switching architecture or changing it by \eg~introducing batch normalization, we need to adapt the hyperparameters -- nodes connected with dashed lines are only semantically similar.
    An apostrophe ($~'~$) denotes evaluations with weakened bias (from $0.9$ to $0.7$), a hat ($~\hat{}~$) architectures with batch normalization, and a tilde in a gray node ($~\tilde{}~$) trainings with the \textit{Sigmoid} instead of the \textit{Factor} mode.
    Analogous nodes for ResNet ($R_3$ and $R_4$) were omitted to improve readability.}
  \label{fig:evaluationcube}
\end{figure}
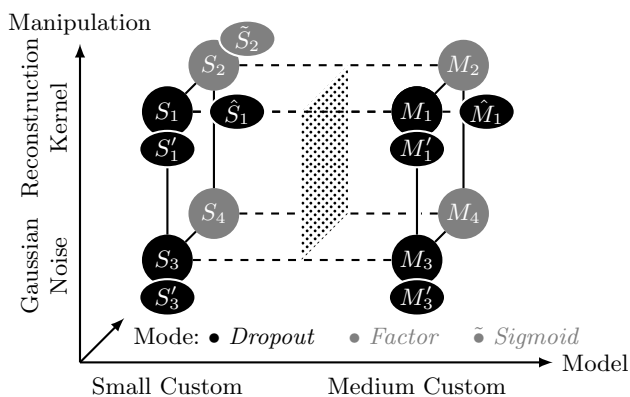

\begin{figure}[htb]
    \begin{tikzpicture}
      \node[inner sep=0] (image) {\includegraphics[width=1.0\linewidth]{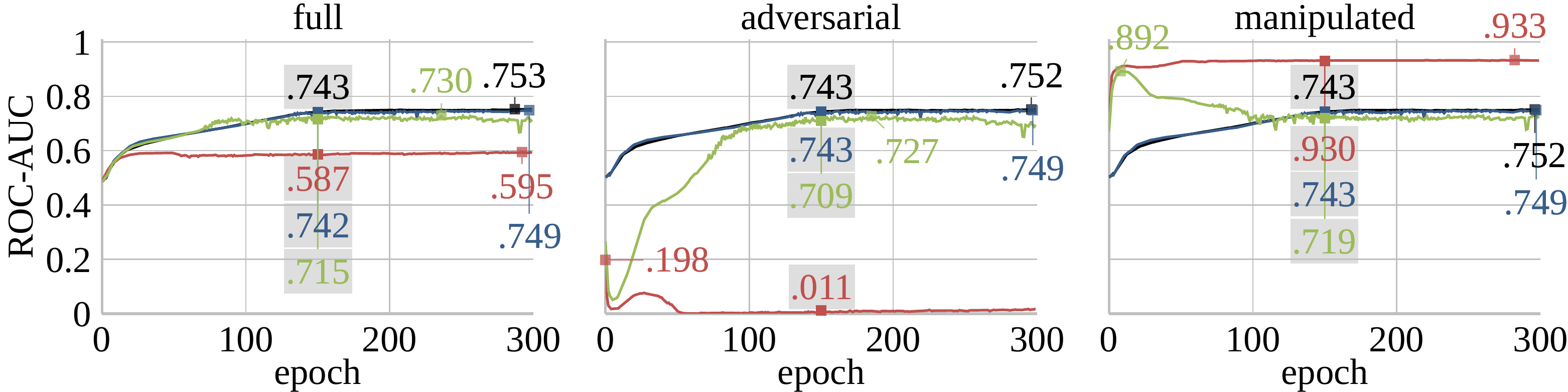}};
      \draw (image.north west) node[anchor=north west,inner sep=0]{$S_1$};
    \end{tikzpicture}

\caption{
        \textit{DecorreLayer} on the small custom architecture. The subplots represent the different test sets: full (unbiased), adversarial, and manipulated (same bias as training data). Color coding denotes whether training data was biased (\color{myred}{\begin{tikzpicture}[square/.style={regular polygon,regular polygon sides=4,inner sep=2pt,ultra thick}]%
          \protect\node at (0,0) [square,draw] (s) {};%
\end{tikzpicture}}\color{black}\,/\,\color{mygreen}{\begin{tikzpicture}[square/.style={regular polygon,regular polygon sides=4,inner sep=2pt,ultra thick}]%
          \protect\node at (0,0) [square,draw,fill] (s) {};%
\end{tikzpicture}}\color{black}) or unbiased (\color{black}{\begin{tikzpicture}[square/.style={regular polygon,regular polygon sides=4,inner sep=2pt,ultra thick}]%
          \protect\node at (0,0) [square,draw] (s) {};%
\end{tikzpicture}}\color{black}\,/\,\color{myblue}{\begin{tikzpicture}[square/.style={regular polygon,regular polygon sides=4,inner sep=2pt,ultra thick}]%
          \protect\node at (0,0) [square,draw,fill] (s) {};%
\end{tikzpicture}}\color{black}) and \textit{DecorreLayer} was enabled or disabled (solid \color{myblue}{\begin{tikzpicture}[square/.style={regular polygon,regular polygon sides=4,inner sep=2pt,ultra thick}]%
          \protect\node at (0,0) [square,draw,fill] (s) {};%
\end{tikzpicture}}\color{black}\,/\,\color{mygreen}{\begin{tikzpicture}[square/.style={regular polygon,regular polygon sides=4,inner sep=2pt,ultra thick}]%
          \protect\node at (0,0) [square,draw,fill] (s) {};%
\end{tikzpicture}}\color{black}~vs.~empty~\color{black}{\begin{tikzpicture}[square/.style={regular polygon,regular polygon sides=4,inner sep=2pt,ultra thick}]%
          \protect\node at (0,0) [square,draw] (s) {};%
\end{tikzpicture}}\color{black}\,/\,\color{myred}{\begin{tikzpicture}[square/.style={regular polygon,regular polygon sides=4,inner sep=2pt,ultra thick}]%
          \protect\node at (0,0) [square,draw] (s) {};%
\end{tikzpicture}}\color{black}~box).}
  \label{fig:eval}
\end{figure}

\begin{table*}[b]
  \footnotesize
  \newcounter{counterCols}
  \setcounter{counterCols}{3}
  \newcounter{counterCols1}\setcounter{counterCols1}{\value{counterCols}}\addtocounter{counterCols}{8}\newcounter{counterCols2}\setcounter{counterCols2}{\value{counterCols}-1}
  \newcolumntype{x}[1]{>{\centering\arraybackslash\hspace{0pt}}p{#1}}
    \setcounter{counterCols}{3}
  \newcounter{counterCols3}\setcounter{counterCols3}{\value{counterCols}}\addtocounter{counterCols}{7}\newcounter{counterCols4}\setcounter{counterCols4}{\value{counterCols}-1}
  \newcounter{counterCols5}\setcounter{counterCols5}{\value{counterCols}}\addtocounter{counterCols}{2}\newcounter{counterCols6}\setcounter{counterCols6}{\value{counterCols}-1}
  \newcounter{counterColsM1}\setcounter{counterColsM1}{\value{counterCols1}-1}
  \newcounter{counterColsM2}\setcounter{counterColsM2}{\value{counterCols1}-2}
  \newcommand{\tablewidth}{0.85cm}
    \caption{Model performance after 150 training epochs. Each column represents a model, with which we performed 4 trainings; when training on biased data, both the unbiased as well as adversarial ROC AUCs are almost always much higher when employing \textit{DecorreLayer} (green filled square) in contrast to not performing any debiasing (red empty square). The respective standard deviations are provided in the supplementary material.}
  \label{tab:performances}
\begin{tabular}{cr x{\tablewidth}x{\tablewidth}x{\tablewidth}x{\tablewidth}x{\tablewidth}x{\tablewidth}x{\tablewidth}x{\tablewidth}}
                                                        & \multirow{2}{*}{\shortstack[l]{\mbox{} \mbox{}                                                                                                                                                                                                                                                                                                                                                                 \\\textbf{Train+Decorr} \\Decorr on \color{myblue}\mycheck\color{black}\,/\,\color{mygreen}\mycheck}} & \multicolumn{8}{c}{Small Custom Architecture} \\\cmidrule(lr){\value{counterCols1}-\value{counterCols2}}
    \textbf{Test}                                       &                                                     & \normalsize$S_1$ & \normalsize$S_1'$ & \normalsize$S_2$ & \normalsize$\tilde{S}_2$ & \normalsize$S_3$ & \normalsize$S_3'$ & \normalsize$S_4$ & \normalsize$\hat{S}_1$  \\ \toprule
    \multirow{4}{*}{\rotatebox{90}{unbiased\,\,\,\,\,\,\,\,\,\,}} & \multirow{2}{*}{unbiased\,} \color{black}\myuncheck & .743             & .743              & .743             & .743                     & .743             & .743              & .743             & .740                    \\
                                                        & \color{myblue}\mycheck                              & .742             & .742              & .739             & .738                     & .742             & .742              & .739             & .749                    \\ \cmidrule(lr){\value{counterColsM1}-\value{counterCols2}}
                                                        & \multirow{2}{*}{biased\,} \color{myred}\myuncheck   & .587             & .663              & .587             & .587                     & .611             & \textbf{.703}     & .611             & \textbf{.656}           \\
                                                        & \color{mygreen}\mycheck                             & \textbf{.715}    & \textbf{.725}     & \textbf{.714}    & \textbf{.706}            & \textbf{.674}    & .701              & \textbf{.686}    & .592                    \\ \cmidrule(lr){\value{counterCols1}-\value{counterCols2}}
                                                        & $p$-value                                    & .001&.001&.001&.003&.075&.810&.051&.203                    \\ \midrule
    \multirow{4}{*}{\rotatebox{90}{adversarial\,\,\,\,\,\,}}      & \multirow{2}{*}{unbiased\,} \color{black}\myuncheck & .743             & .743              & .743             & .743                     & .743             & .743              & .743             & .742                    \\
                                                        & \color{myblue}\mycheck                              & .743             & .743              & .740             & .739                     & .743             & .743              & .740             & .750                    \\ \cmidrule(lr){\value{counterColsM1}-\value{counterCols2}}
                                                        & \multirow{2}{*}{biased\,} \color{myred}\myuncheck   & .011             & .286              & .011             & .011                     & .011             & .314              & .011             & .293                    \\
                                                        & \color{mygreen}\mycheck                             & \textbf{.709}    & \textbf{.744}     & \textbf{.644}    & \textbf{.607}            & \textbf{.655}    & \textbf{.596}     & \textbf{.622}    & \textbf{.543}           \\ \cmidrule(lr){\value{counterCols1}-\value{counterCols2}}
                                                        & $p$-value                                    &.000&.000&.000&.000&.000&.000&.000&.107     \\\bottomrule
\end{tabular}%
\\\mbox{}\\\mbox{}\\%
\begin{tabular}{cr x{\tablewidth}x{\tablewidth}x{\tablewidth}x{\tablewidth}x{\tablewidth}x{\tablewidth}x{\tablewidth} x{\tablewidth}x{\tablewidth}}
                                                        & \multirow{2}{*}{\shortstack[l]{\mbox{} \mbox{}                                                                                                                                                                                                                                                                                                                                                                 \\\textbf{Train+Decorr} \\Decorr on \color{myblue}\mycheck\color{black}\,/\,\color{mygreen}\mycheck}} &  \multicolumn{7}{c}{Medium Custom Architecture} & \multicolumn{2}{c}{ResNet-18}  \\\cmidrule(lr){\value{counterCols3}-\value{counterCols4}}\cmidrule(lr){\value{counterCols5}-\value{counterCols6}}
    \textbf{Test}                                       &                                                     & \normalsize$M_1$ & \normalsize$M_1'$ & \normalsize$M_2$ & \normalsize$M_3$ & \normalsize$M_3'$ & \normalsize$M_4$ & \normalsize$\hat{M}_1$ & \normalsize$R_3$ & \normalsize$R_4$ \\ \toprule
    \multirow{4}{*}{\rotatebox{90}{unbiased\,\,\,\,\,\,\,\,\,\,}} & \multirow{2}{*}{unbiased\,} \color{black}\myuncheck & .731             & .731              & .731             & .731             & .731              & .731             & .730                   & .740             & .740             \\
                                                        & \color{myblue}\mycheck                              & .703             & .703              & .752             & .703             & .703              & .752             & .710                   & .557             & .614             \\ \cmidrule(lr){\value{counterColsM1}-\value{counterCols6}}
                                                        & \multirow{2}{*}{biased\,} \color{myred}\myuncheck   & .554             & .634              & .554             & .574             & .670              & .574             & \textbf{.626}          & \textbf{.645}    & .645             \\
                                                        & \color{mygreen}\mycheck                             & \textbf{.713}    & \textbf{.677}     & \textbf{.554}    & \textbf{.667}    & \textbf{.711}     & \textbf{.726}    & .617                   & .570             & \textbf{.699}    \\ \cmidrule(lr){\value{counterCols1}-\value{counterCols6}}
                                                        & $p$-value                                    &.006&.343&.998&.065&.357&.016&.846&.007&.041             \\ \midrule
    \multirow{4}{*}{\rotatebox{90}{adversarial\,\,\,\,\,\,}}      & \multirow{2}{*}{unbiased\,} \color{black}\myuncheck & .729             & .729              & .729             & .729             & .729              & .729             & .729                   & .739             & .739             \\
                                                        & \color{myblue}\mycheck                              & .707             & .707              & .747             & .707             & .707              & .747             & .710                   & .558             & .618             \\ \cmidrule(lr){\value{counterColsM1}-\value{counterCols6}}
                                                        & \multirow{2}{*}{biased\,} \color{myred}\myuncheck   & .240             & .343              & \textbf{.240}    & .492             & .529              & .492             & .224                   & .016             & .016             \\
                                                        & \color{mygreen}\mycheck                             & \textbf{.749}    & \textbf{.725}     & .204             & \textbf{.667}    & \textbf{.711}     & \textbf{.719}    & \textbf{.272}          & \textbf{.711}    & \textbf{.652}    \\ \cmidrule(lr){\value{counterCols1}-\value{counterCols6}}
                                                        & $p$-value                                    &.013&.020&.802&.119&.161&.033&.679&.000&.000                            \\\bottomrule
\end{tabular}%
\end{table*}

\newpage
\section{Evaluation and Discussion}
\label{sec:evaluation}
The mean ROC-AUC of a 5-fold cross validation serves as the main evaluation metric; Fig.~\ref{fig:evaluationcube} visualizes and explains the test setups represented by abbreviations such as $S_1$. The results are summarized in Table \ref{tab:performances} as the ROC-AUCs after half of the (typically over-)allocated training time of 300 epochs.

In Fig.~\ref{fig:eval} ($S_1$), we demonstrate the inflated score one would see when \eg~training and testing the small custom architecture on data with reconstruction kernel bias (ROC-AUC $.930$), compared to the dramatic impact on true ($.587$ ROC-AUC) and especially adversarial performance ($.011$ ROC-AUC). \textit{DecorreLayer} minimizes this issue, achieving scores within $.035$ ROC-AUC of the baseline using the recommended \textit{Dropout} mode of the Filter Unit.

\textit{DecorreLayer} also effectively decorrelates when applied to weaker reconstruction kernel bias ($S_1'$) and/or incorporated into the medium custom architecture ($M_1$, $M_1'$). Further tests of AWGN added to most images of ill patients ($S_3$, $S_3'$, $S_4$, $M_3$, $M_3', M_4$) show that hyperparameters \eg~gained from another type of bias can serve as a reasonable basis for new trainings. The performance of \textit{Factor} ($S_2$, $M_2$) or \textit{Sigmoid} ($\tilde{S}_2$) Filter Unit modes is typically worse, and their decorrelation effect is less sustained -- when examining the full plots, (especially adversarial) performance tends to drop again in favor of re-learning the bias.

Applying \textit{DecorreLayer} to the comparatively huge ResNet-18 did not work as easily ($R_3$); further investigation would be necessary to determine if \textit{DecorreLayer} can work consistently on ResNet-18 with dataset sizes common in medical applications. We decorrelated a training on AWGN manipulated data using the \textit{Factor} mode ($R_4$) at the cost of degraded performance if no bias had been present. However, we observed promising results when performing additional testing with ResNet-18 on CIFAR-10 \cite{Krizhevsky2009LearningML} (cf.~supplementary material, Figs.~1 and 2).
Also, \textit{DecorreLayer} is currently incompatible with batch normalization; we argue that the mean and variance estimates of batch normalization might fluctuate significantly due to the changing number of filtered features, hence creating conflicts.

Having determined appropriate hyperparameters for \textit{DecorreLayer} using the proposed Histogram of Correlations, 
it has very little negative -- and sometimes even a positive -- impact on training stability. For instance, we observed occasional stability issues causing divergence to 0.5 ROC-AUC when training our medium custom architecture on biased data; in this case, \textit{DecorreLayer} acted in a regularizing fashion like a regular dropout layer \cite{Srivastava2014DropoutAS}. Furthermore, we still consider access to a small unbiased data set vital for real-world application: to determine suitable hyperparameters and gain interpretable testing results.

\clearpage
\section{Conclusion and Outlook}
To the best of our knowledge, \textit{DecorreLayer} is the first method to debias DL models while training on data containing unknown biases: we teach models not to use bias present in training data as long as the confounder is also observable in image areas that are not directly related to the task at hand. While currently working less reliably on larger architectures, as well as requiring hyperparameter values carefully adjusted using the proposed tools, in our experiments on CT lung emphysema classification \textit{DecorreLayer} was able to immensely boost generalization performance under adverse conditions of both artificially created and intentionally sampled technical variation -- often near-perfectly recovering the baseline performance while training on heavily biased data.

To achieve this, instances of \textit{DecorreLayer} are inserted into the model pipeline, they analyze how the model perceives control regions in addition to the regions of interest. It does not entail \eg~an additional loss term, but integrates into the architecture as a smart filtering/regularization layer. During training, computational requirements increase somewhat due to an additional forward pass for the control region. In exchange, \textit{DecorreLayer} introduces no trainable parameters since it re-uses the feature extraction of the main model and does not create any load when performance matters most, \ie~during inference -- the original architecture can perform its task without extra help.
We argue that slightly relaxing the correctness of the gradient calculation can provide new ways to steer training into the intended direction.

Future work directly expanding on \textit{DecorreLayer} could learn/dynamically calculate Filter Unit hyperparameters, or investigate new Correlation Units: \textsc{a)} applying other existing measures of dependence like the Hilbert-Schmidt independence criterion \cite{Gretton2007AKS} or mutual information \cite{Shannon1948AMT} and, \textsc{b)} learning an arbitrary correlation function, which could circumvent the risk of models bypassing correlation checks by considering earlier feature maps (or even unprocessed CRs) to compare ROI features against. Future work also needs to investigate how well the approach generalizes to various modalities as well as other types of bias.
We expect long-term developments to head towards more directly informing the main architecture of what it did wrong when learning biased features: for instance by combining techniques limited to known bias like adversarial debiasing with un- or self-supervised techniques such as VAEs \cite{Kingma2014AutoEncodingVB}, or SimCLR(v2) \cite{Chen2020ASF,Chen2020BigSM} and BYOL \cite{Grill2020BootstrapYO} to generate representations of technical variation.

\clearpage % to force floats being flushed before references

\bibliographystyle{splncs04}
\bibliography{refs}
\end{document}